\pdfoutput=1

\documentclass[11pt]{article}

\usepackage{acl}

\usepackage{times}
\usepackage{latexsym}
\usepackage{amsfonts}
\usepackage{geometry}
\usepackage{array}
\usepackage{float} 
\usepackage{tablefootnote}
\usepackage[T1]{fontenc}
\usepackage{pgfplots}
\pgfplotsset{compat=1.18}
\usepackage{booktabs}
\usepackage{xcolor}
\usepackage{listings}
\usepackage[utf8]{inputenc}

\usepackage{microtype}

\usepackage{inconsolata}
\usepackage{amsmath}
\usepackage{bm}
\usepackage{algorithm}
\usepackage{algpseudocode}
\usepackage{graphicx}
\usepackage{svg}
\usepackage{multirow}
\usepackage{booktabs}
\usepackage{tcolorbox}
\tcbuselibrary{breakable}
\usepackage{xspace}
\usepackage{CJKutf8}
\usepackage{tabularx}
\usepackage{stfloats}

\title{RAISE: Reinforced Adaptive Instruction Selection For Large Language Models}

\author{
\textbf{Qingsong Lv}\textsuperscript{*1},
\textbf{Yangning Li}\textsuperscript{*12},
\textbf{Zihua Lan}\textsuperscript{1},
\textbf{Zishan Xu}\textsuperscript{1},
\textbf{Jiwei Tang}\textsuperscript{1},
\textbf{Tingwei Lu}\textsuperscript{1} \\
\textbf{Yinghui Li}\textsuperscript{1},
\textbf{Wenhao Jiang}\textsuperscript{5},
\textbf{Hong-Gee Kim}\textsuperscript{4},
\textbf{Hai-Tao Zheng}\textsuperscript{\dag12},
\textbf{Philip S. Yu}\textsuperscript{3} \\
\\
\textsuperscript{1}Shenzhen International Graduate School, Tsinghua University \\ \textsuperscript{2}Peng Cheng Laboratory \textsuperscript{3}University of Illinois Chicago \textsuperscript{4}Seoul National University \\
\textsuperscript{5}Guangdong Laboratory of Artificial Intelligence and Digital Economy (SZ)   \\
\texttt{\{lqs23,yn-li23\}@mails.tsinghua.edu.cn} \\
\texttt{zheng.haitao@sz.tsinghua.edu.cn}
}

\begin{document}
\maketitle

\renewcommand{\thefootnote}{}
\footnotetext{* Equal contribution.}
\renewcommand{\thefootnote}{\arabic{footnote}}
\renewcommand{\thefootnote}{}
\footnotetext{\dag Corresponding author.}
\renewcommand{\thefootnote}{\arabic{footnote}}

\begin{abstract}
Instruction tuning of large language models (LLMs) benefits more from a handful of high-quality examples than from hordes of low-quality ones. 
Existing selection methods typically rely on static, heuristic quality scores and are executed only once before training. 
Consequently, they neither adapt to the changing state of the model nor target downstream objectives, leaving substantial room for optimization. 
We propose RAISE (\textbf{R}einforced \textbf{A}daptive \textbf{I}nstruction \textbf{SE}lection), a \emph{dynamic}, \emph{task-driven} framework that integrates selection into every training step. 
At each step, RAISE estimates the expected contribution of each candidate instruction to task performance and admits only the most helpful. 
By modeling this process as sequential decision making, we optimize the selector with reinforcement learning, yielding an interpretable policy specialized for the target task. 
Extensive experiments show that RAISE reaches comparable or better results than full-data training while updating only 1\% of the steps, demonstrating both high efficacy and significant computational savings.
\end{abstract}


\section{Introduction}
Large Language Models (LLMs) have achieved remarkable progress in recent years, demonstrating exceptional capabilities in general language understanding \cite{liu2023mmc,chen2024hytrel} and generation \cite{openai2023optimizing,achiam2023gpt,liu2024large,sun2024itd,tang2024perception,han2024length,zhang2025not,yuan2025probabilistic,lv2026data,tang2026comi,tang2026read}. A critical factor enabling these advancements is instruction fine-tuning \cite{wei2021finetuned,longpre2023flan,chung2024scaling,tang2025gmsa}, a process that aligns pretrained models with human intentions by training them on task-specific instructions. While existing efforts predominantly focus on scaling instruction datasets \cite{khashabi2020unifiedqa,ye2021crossfit,wang2022super,han2025attributes} to improve model performance, recent studies highlight that data quality often outweighs sheer quantity \cite{zhou2024lima}. This underscores the need for principled methods to identify instruction subsets that maximally enhance model capabilities.

Current instruction selection approaches typically rely on heuristic quality metrics (eg. grammatical correctness, clarity, lexical diversity, etc.) to filter low-quality instructions before training \cite{cao2023instruction,li2023quantity,chen2023alpagasus,xia2024less,pan2024g}. These methods face three main issues: (i) They use a one-time static selection before training , which does not adapt to a model’s evolving data needs during training; (ii) Their heuristic metrics are prone to cognitive bias and oversimplify the continuous nature of data quality; (iii) They are task-agnostic, failing to align instruction selection with specific task objectives.

Considering a dynamic, task-aware approach to instruction selection, we introduce the concept of an instruction’s \textbf{dynamic value}—its impact on the final model performance when used for gradient updates at time step \(t\) (we put detailed descriptions of dynamic value and example in Appendix~\ref{app:dynamic_value}). This dynamic value, which depends on both the training step and the task objective, serves as a quality measure that replaces fixed heuristic metrics and provides strong interpretability. Dynamic instruction selection can be modeled as a sequential decision-making process aiming to maximize the model's performance after \(T\) steps. Obviously, the optimal selection strategy is to select those instructions that have the most dynamic value at each step.

Based on this idea, we propose \textbf{RAISE} (\textbf{R}einforced \textbf{A}daptive \textbf{I}nstruction \textbf{SE}lection), a dynamic, non-heuristic, task-driven instruction selection framework. At its core is an \textbf{acquisition function}—a trainable MLP (sample-wise scorer) that estimates the \textbf{dynamic value} of each instruction and is optimized to maximize the final model performance. By leveraging a fully trained acquisition function to guide instruction selection, RAISE consistently produces high-performing models. Moreover, its task-driven design allows the acquisition function to be flexibly adapted to various tasks through adjustments in the validation set and performance metrics. Due to the sequential decision-making nature of RAISE, we employ reinforcement learning (RL) \cite{bellman1966dynamic,mnih2015human} to optimize the acquisition function, treating each training process of LLM as an episode in the RL setting.

\begin{figure*}
    \centering
    \scalebox{1}{
    \includegraphics[width=1\linewidth]{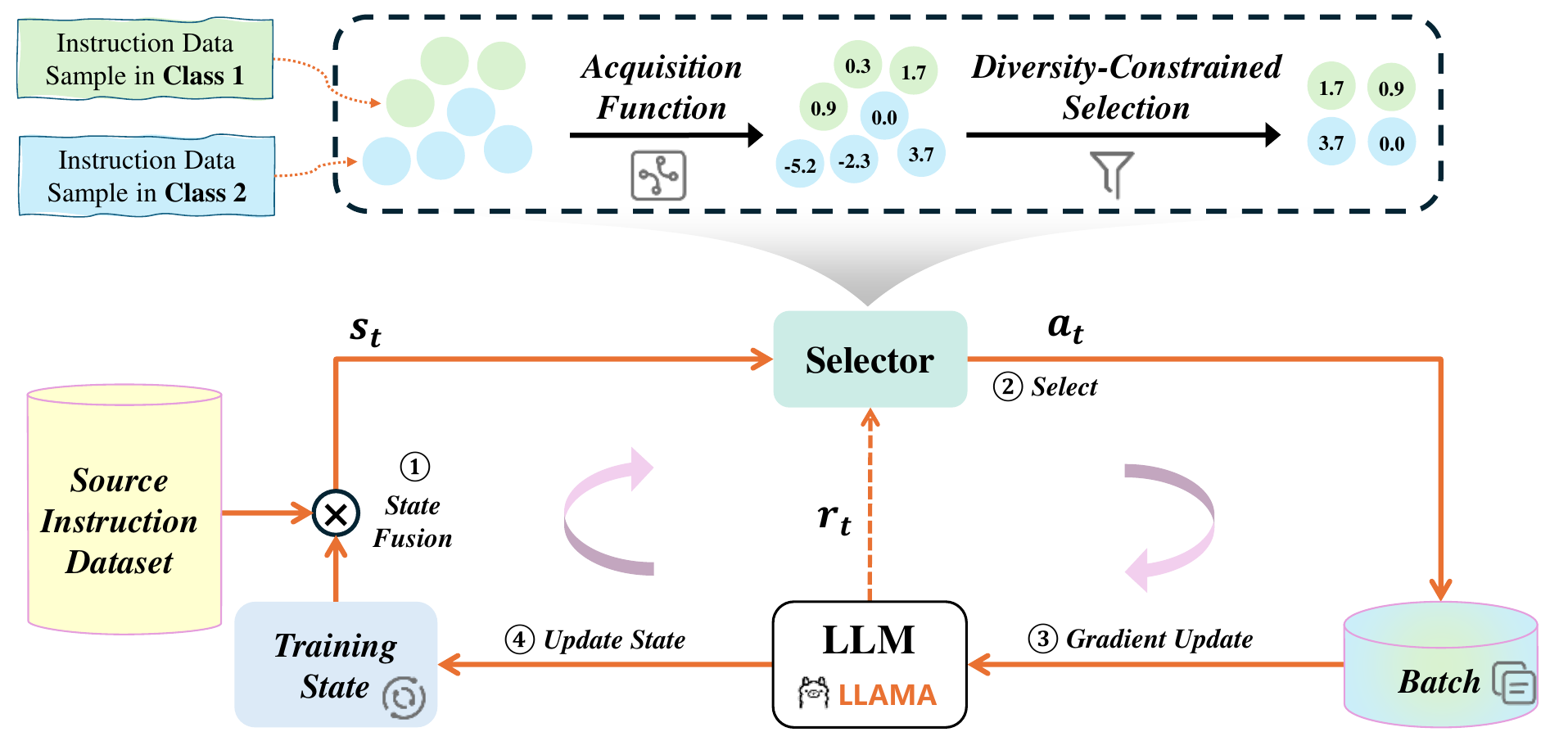}}
    \caption{Overview of the RAISE framework, illustrating the training process of LLM at step $t$:
    (1) The \textbf{source instruction dataset} and \textbf{current training state} are fused to construct $s_t$, which encodes both data features and training progress. (2) The \textbf{Selector}, guided by the \textbf{acquisition function} (a trainable scorer), takes $s_t$ as input and selects a batch of instruction data. (3) This selected batch is used to update LLM, resulting in performance improvement $r_t$. (4) Finally, the updated training state forms $s_{t+1}$, serving as input for the next step. }
    \label{fig:intro}
\end{figure*}

To promote diversity , RAISE groups instructions into multiple classes via K-means \cite{macqueen1967some} and ensures balanced sampling from each group in every training batch. These classes constitute diversity constraint. Figure~\ref{fig:intro} shows the framework for RAISE, which considers both the score of the acquisition function and the diversity constraint when selecting instructions.

In summary, our contributions are as follows:
\begin{itemize}
\item We design a \textbf{task-objective-driven acquisition function} to estimate the dynamic value of each instruction based on its expected impact on the final model performance, eliminating the need for heuristic quality metrics.
\item We propose \textbf{RAISE}, a dynamic instruction selection framework that adaptively selects instructions during training based on their dynamic value, enabling the model to meet changing data requirements during training.
\item Through extensive experiments, we demonstrate the effectiveness of our approach and provide analysis highlighting the potential of dynamic instruction selection for future advancements in instruction fine-tuning.
\end{itemize}

\section{Related Work}
\subsection{Instruction Selection}
Instruction selection focuses on identifying a subset of instructions from a fine-tuning dataset that maximize model performance, rather than training on the entire dataset. Recent studies emphasize that carefully selected subsets can often outperform full-dataset training \cite{cao2023instruction,li2023quantity,xia2024less}, underscoring the importance of effective selection strategies. The shift from dataset scaling to quality-focused selection highlights the need for principled methods to prioritize high-utility instructions.

Many methods have been proposed for instruction selection. \textbf{IFD} \cite{li-etal-2024-quantity} introduces an Instruction Following Difficulty metric to assess instruction complexity and select appropriate samples. \textbf{AlpaGasus} \cite{chen2024alpagasus} uses GPT-4 \cite{achiam2023gpt} to score instruction-response pairs, filtering low-quality samples and improving training efficiency. \textbf{DEITA} \cite{liu2024what} combines complexity and quality scores to optimize instruction selection, balancing diversity and data quality.

However, these approaches share a common limitation: \textbf{static selection}. Once the subset is chosen, it remains fixed throughout training, failing to adapt to the model’s evolving data preferences. In contrast, our proposed RAISE framework dynamically selects instructions at each training step based on their dynamic value, enabling adaptive learning that aligns with the model’s changing needs.




\subsection{Self-Paced Learning}
Self-Paced Learning (SPL) \cite{kumar2010self} represents a prominent curriculum learning paradigm \cite{bengio2009curriculum} that dynamically selects training samples based on their difficulty levels. Unlike static curriculum approaches, SPL employs an adaptive weighting mechanism where easier samples are prioritized in early training stages while harder ones are progressively incorporated. 

The core mechanism involves jointly optimizing model parameters and sample selection through a bi-level objective: while the model learns to minimize task loss, the sample selector determines optimal inclusion thresholds based on current loss values. This loss-driven thresholding strategy has proven effective in improving convergence robustness across various domains \cite{wang2021survey}.

However, this loss-driven approach introduces critical limitations for instruction tuning: (i) Loss values often poorly reflect task-specific metrics (e.g., accuracy, BLEU); (ii) The rigid easy-to-hard progression may discard valuable hard samples; (iii) Its single optimization objective cannot adapt to diverse task requirements. RAISE addresses these issues by replacing loss with task-aware dynamic value estimation and introducing diversity constraint through clustered sampling, enabling both task-aware selection and adaptive learning.

\section{Method}
In this section, we describe our method for dynamic instruction selection. A learnable acquisition function is trained to estimate the dynamic value of each instruction, ensuring adaptive and diversity-aware selection throughout the training process.

We formally define the problem of dynamic instruction selection (\S~\ref{sec:problem}), introduce the training framework of our selection policy (\S~\ref{sec:training_framework}), and describe the state fusion mechanism that combines training state and data features (\S~\ref{sec:state_fusion}). We then present the instruction selection algorithm (\S~\ref{sec:select_algo}) and the policy optimization algorithm for improving the selection policy (\S~\ref{sec:policy_optim}).

\subsection{Problem Statements}
\label{sec:problem}
Given an instruction dataset $\mathcal{D} = \{d_i\}_{i=1}^{N}$, our goal is to dynamically select a subset $\mathcal{D}_t$ at each training step $t$ to maximize the performance $\mathcal{P}$ of the model updated at final step $T$. $\mathcal{P}$ is defined as a performance metric related to downstream tasks, which includes a validation set $\mathcal{D}_\text{val}$ visible during the training process and an evaluation metric (e.g., -loss/accuracy/rouge on $\mathcal{D}_\text{val}$). The optimal selection policy $\pi^*$ can be formulated as:
\begin{equation} 
\pi^* = \arg\max_{\pi} \mathcal{P}(\mathcal{M}_T[\mathcal{D}, \pi], \mathcal{D}_\text{val}), 
\end{equation}
where, $\mathcal{M}_T[\mathcal{D}, \pi]$ represents the model updated at step $T$. 
For simplicity, in the following content, we denote $\mathcal{P}(\,\cdot\,, \mathcal{D}_\text{val})$ as $\mathcal{P}(\,\cdot\,)$ and $\mathcal{M}_t[\mathcal{D}, \pi]$ as $\mathcal{M}_t$.

\subsection{Training Framework of Selection Policy}
\label{sec:training_framework}
Dynamic instruction selection can be formulated as a sequential decision-making process. Specifically, at each training step $t$, the selection policy $\pi$ determines a subset $\mathcal{D}_t$ from the dataset $\mathcal{D}$ to update the model $\mathcal{M}_{t-1}$. This process can be modeled as a Markov Decision Process (MDP) \cite{bellman1966dynamic,puterman2014markov} consisting of:
\begin{itemize}
\item \textbf{State ($S_t$)}: The state at step $t$, represents all available information, building from the current training state and $\mathcal{D}$ by \textbf{State Fusion}.
\item \textbf{Action ($A_t$)}: The action is the selected batch data $\mathcal{D}_t$ from $\mathcal{D}$ according to the policy $\pi$, i.e., $A_t = \mathcal{D}_t = \pi(S_t)$.
\item \textbf{Reward ($R_t$)}: The reward is based on the performance improvement after using $\mathcal{D}_t$ to update the model, i.e., $R_t = \mathcal{P}(\mathcal{M}_t) - \mathcal{P}(\mathcal{M}_{t-1})$. 
\end{itemize}

Once the subset $\mathcal{D}_t$ is selected, it is used to update the model $\mathcal{M}_{t-1}$, resulting in the new model $\mathcal{M}_t$ and an updated state $S_{t+1}$. The goal of training is to maximize the cumulative reward, which reflects the final model performance $\mathcal{P}(\mathcal{M}_T)$.

In this framework, the selection policy $\pi$ consists of a learnable acquisition function $\mathcal{F}$ and a diversity constraint $\mathcal{C}$. Only $\mathcal{F}$ is trainable, so optimizing $\pi$ is equivalent to optimizing $\mathcal{F}$. 

\subsection{State Fusion}
\label{sec:state_fusion}
In dynamic instruction selection, \textbf{State Fusion} combines the current training state with original instruction features to form a comprehensive representation for the acquisition function. Specifically, we denote the fused state as $d'=\mathcal{H}(d,\mathcal{M}_{t-1},t)$, where $d$ is the instruction sample and $\mathcal{H}$ is the fusion function. The fusion of state involves 4 components:
\begin{itemize}
\item \textbf{Stage State} ($\mathcal{H}_{\text{stage}}$): This component captures the model’s current training progress, including $\mathcal{M}_{t-1}$ and $t$. Formally: 
\begin{equation}
\mathcal{H}_{stage}(\mathcal{M}_{t-1}, t) = \left[ \mathcal{P}_{t-1}, \dfrac{t}{T} \right]
\end{equation}
\item \textbf{Instruction-Difficulty State} ($\mathcal{H}_{\text{diff}}$): To represent the complexity of each instruction, we collect $log P(y|x), log P(y)$, and the lengths of the prompt and its response. To ensure efficiency, they are precomputed using the auxiliary model\footnote{We use Llama-3.1-8B-Instruct as the auxiliary model to preprocess instruction-difficulty state and instruction embeddings.}. Formally:
\begin{equation}
\begin{split}
\mathcal{H}_{\text{diff}}(d) = \left[\text{len}(x), \text{len}(y),\right. \\
\left. \log P(y|x), \log P(y) \right]
\end{split}
\end{equation}
\item \textbf{Instruction-Semantic State} ($\mathcal{H}_{\text{sem}}$): This component encodes the semantic information of the instruction. We compute the embedding vector $E(d)$ with the auxiliary model, followed by a pooling layer:
\begin{equation}
\mathcal{H}_{\text{sem}}(d) = \left [\text{Pool}(E(d)) \right ]
\end{equation}
\item \textbf{Instruction-Availability State} ($\mathcal{H}_{\text{avail}}$): We record the number of times $\nu(d)$ an instruction has already been selected during training, helping the acquisition function avoid excessive repetition of the same instruction:
\begin{equation}
\mathcal{H}_{\text{avail}}(d) = \left [ \nu(d) \right ]
\end{equation}
\end{itemize}
By concatenating these 4 components, we obtain the fused state:
\begin{equation}
\begin{split}
\mathcal{H}(d, \mathcal{M}_{t-1}, t) = \left [\mathcal{H}_{\text{stage}}(\mathcal{M}_{t-1}, t), \right. \\
\left. \mathcal{H}_{\text{dif}}(d), \mathcal{H}_{\text{sem}}(d), \mathcal{H}_{\text{avail}}(d)\right ]    
\end{split}
\end{equation}

\subsection{Instruction Selection Algorithm}
\label{sec:select_algo}
\begin{algorithm}
\small
\caption{Dynamic Instruction Selection with Diversity Constraint}
\label{alg:dynamic_selection}
\begin{algorithmic}[1]
\State \textbf{Input:} Training dataset \( \mathcal{D} \), LLM \( \mathcal{M}_{t-1} \), Batch size \( B \), Acquisition function \( \mathcal{F} \), Diversity constraint (classes) \( \mathcal{C} \) and Fusion function \( \mathcal{H} \)
\State \textbf{Output:} Selected subset of \( B \) samples
\State \( C \gets |\mathcal{C}| \), \( b \gets \frac{B}{C} \)
\State Initialize \( S_t \gets \emptyset, s \gets \emptyset \)
\For{\( d_j \in \mathcal{D} \)}
    \State \( d'_j \gets \mathcal{H}(d_j, \mathcal{M}_{t-1}, t) \)
    \State \( S_t \gets S_t \cup \{d'_j\} \)
    \State \( s_j \gets \mathcal{F}(d'_j) \)  \Comment{Dynamic value of \(d_j\)}
\EndFor
\For{\( \mathcal{C}_i \in \mathcal{C} \)} \Comment{Divide \( S_t \) into \( C \) classes}
    \State \( S_{t,i} \gets \emptyset \)
    \For{ \( d_j \in \mathcal{C}_i \)}
        \State \( S_{t,i} \gets S_{t,i} \cup d'_j \)
    \EndFor
\EndFor 
\For{\( i = 1, 2, \dots, C \)}
    \State \( \pi(S_{t, i}) \gets \operatorname{arg\,top}_b \{s_j \mid d'_j \in S_{t, i}\} \)
\EndFor 
\State \( \pi(S_t) \gets \bigcup_{i=1}^C \pi(S_{t, i}) \)
\State \Return \( \pi(S_t) \)
\end{algorithmic}
\end{algorithm}
Algorithm~\ref{alg:dynamic_selection} presents the instruction selection process with diversity constraint at training step $t$. We first apply the fusion function $\mathcal{H}$ to incorporate training state into each instruction $d_j$. The acquisition function $\mathcal{F}$ then scores the fused instructions, and a diversity constraint $\mathcal{C} = \{\mathcal{C}_1,\dots,\mathcal{C}_C\}$ (each $\mathcal{C}_i$ represents a class) ensures balanced coverage of heterogeneous instruction types. Specifically, we select the top-$b$ instructions (based on $\mathcal{F}$) from each class, and their union forms the final training subset $\mathcal{D}_t$. This selected batch is then used to update LLM, and the process repeats at the next training step.

\subsection{Policy Optimization Algorithm}
\label{sec:policy_optim}
To train the selection policy $\pi$, we adopt PPO \cite{schulman2017proximal}, where the acquisition function $\mathcal{F}_{\theta}$ acts as \textbf{Actor}, and $V_{\phi}$ serves as \textbf{Critic}. 

\paragraph{Advantage Estimation.}
To stabilize training and improve generalization, we employ Generalized Advantage Estimator (GAE) \cite{schulman2015high} for advantage computation:
\begin{equation}
\delta_t \;=\; R_t + \gamma\,V_{\phi}(S_{t+1}) \;-\; V_{\phi}(S_t),
\end{equation}
\begin{equation}
\text{Adv}_t \;=\; \sum_{l=0}^{T-t-1}\; (\gamma\,\lambda)^l\;\delta_{t+l},
\end{equation}
\begin{equation}
G_t \;=\; V_{\phi}(S_t)\;+\;\text{Adv}_t,
\end{equation}
where $\gamma$ is the discount factor and $\lambda$ is the GAE parameter, $\text{Adv}_t$ and $G_t$ is advantage and return respectively.

\paragraph{Importance Sampling with Diversity Constraint.}
Under the diversity-constrained selection, the importance sampling ratio is computed on a per-class basis. Let $\{ \mathcal{C}_1,\dots,\mathcal{C}_C \}$ be the class used in instruction selection, and define:
\begin{equation}
\begin{split}
p_{\text{new}, i}(d'_j) \;=\;
\frac{\exp\bigl(\mathcal{F}_{\theta}(d'_j)\bigr)}{\sum_{d'_k \in \mathcal{C}_i} \exp\bigl(\mathcal{F}_{\theta}(d'_k)\bigr)}, \\
p_{\text{old}, i}(d'_j) \;=\;
\frac{\exp\bigl(\mathcal{F}_{\theta_{\text{old}}}(d'_j)\bigr)}{\sum_{d'_k \in \mathcal{C}_i} \exp\bigl(\mathcal{F}_{\theta_{\text{old}}}(d'_k)\bigr)},
\end{split}
\end{equation}
Then, the overall ratio for a selected batch is:
\begin{equation}
\hat{r}_t 
\;=\;
\prod_{i=1}^{C}\;
\prod_{d'_j \in \,\pi(S_{t,i})}
\;\frac{p_{\text{new}, i}(d'_j)}{p_{\text{old}, i}(d'_j)},
\end{equation}
where $\pi(S_{t,i})$ denotes the top-$b$ samples chosen from the $i$-th class at step $t$.

\paragraph{Loss Functions.}
Following PPO, we optimize both the actor and critic losses. The actor loss is given by:
\begin{equation}
\begin{split}
\mathcal{L}^{\text{actor}}
\;=\;
-\,\mathbb{E}_t\,
\Bigl[
\min\Bigl(
\hat{r}_t\;\text{Adv}_t,\; \\
\text{clip}\bigl(\hat{r}_t,1-\epsilon,1+\epsilon\bigr)\;\text{Adv}_t
\Bigr)
\Bigr],
\end{split}
\label{loss_actor}
\end{equation}
where $\epsilon$ is the clipping parameter ($\epsilon=0.2$). The critic loss is simply a mean-squared error:
\begin{equation}
\mathcal{L}^{\text{critic}}
\;=\;
\mathbb{E}_t\,
\Bigl[
\bigl(
V_{\phi}(S_t)
\;-\;
G_t
\bigr)^2
\Bigr].
\label{loss_critic}
\end{equation}
\paragraph{Training Framework.}
We run $K$ rounds of PPO training. In each round, the LLM is trained for $T$ steps following the current policy $\pi$, with data (i.e., states, actions, rewards) being collected. We then use these collected samples to update the actor $\mathcal{F}_\theta$ and the critic $V_\phi$ via the aforementioned PPO objective. Iterating this process over $K$ rounds gradually refines the acquisition function $\mathcal{F}_\theta$, ultimately yielding a strong policy for dynamic instruction selection. Detailed training process is in Appendix~\ref{app:ppo}.

\section{Experiments}

\subsection{Experimental Setup}
\paragraph{Training Dataset.} We use Alpaca-52K \cite{alpaca} as our instruction fine-tuning dataset, which contains 52,000 multi-domain instruction-response pairs spanning tasks such as question answering, text generation, translation and so on\footnote{We use the Alpaca version from \url{https://huggingface.co/datasets/yahma/alpaca-cleaned}.}.

\paragraph{Evaluation Datasets.} We evaluate on four benchmarks: \textbf{MMLU}, \textbf{ARC (Challenge)} (\textbf{ARC-C}), \textbf{CommonsenseQA} (\textbf{ComQA}), and \textbf{GSM8K}. See Appendix~\ref{app:eval_detail} for detailed settings of them.

\subsection{Baselines}
We employ multiple baselines to compare with RAISE. The simplest one is \textbf{RAND}, which randomly samples a subset of instructions from the full training set. We report the average performance over 5 independent random samplings. We also compare against other established methods, such as \textbf{IFD}, \textbf{DEITA} and \textbf{AlpaGasus}. In addition, we design a dynamic selection variant for SPL, termed \textbf{SSPL} (see Appendix~\ref{app:sspl}). 

To ensure a fair comparison between static and dynamic selection methods, we match the total number of update steps across all approaches. Concretely, for static methods, we first pick $1\%$ of the full training set as a fixed subset and then train the model for 3 epochs. For dynamic methods, we set $\texttt{max\_steps}$ to match the total number of update steps in the static setting, thereby enforcing an equivalent amount of training.

\subsection{Main Results}
\begin{table*}[hb]
    \centering
    \begin{tabular}{c c c c| c c c}
    \toprule
    \textbf{Model} & \textbf{DATA} & \textbf{Avg.Q} & \textbf{Avg.} & \textbf{MMLU} & \textbf{ARC-C} & \textbf{ComQA} \\
    \midrule
    \multirow{7}{*}{Llama-3.2-3B}
    & 0\%            & -100\%    & 52.67 & 51.66 & 42.15 & 64.21 \\
    & 100\%          & 0\%       & 54.32 & 52.76 & 43.77 & \textbf{66.42} \\
    \cline{2-7}
    & RAND           & -7.33\%   & 54.20 & 52.86 & 42.32 & 65.11 \\
    & IFD            & +25.06\%  & 54.73 & 52.66 & 46.42 & 65.11 \\
    & DEITA          & -16.44\%  & 54.05 & 51.90 & 44.88 & 65.36 \\
    & AlpaGasus      & -72.14\%  & 53.13 & 52.30 & 44.11 & 62.98 \\
    & SSPL           & -196.75\% & 51.08 & 50.11 & 41.64 & 61.51 \\
    & \textbf{RAISE(Ours)} & \textbf{+70.35\%} & \textbf{55.47} & \textbf{54.64} & \textbf{46.59} & 65.19 \\
    \midrule
    \multirow{7}{*}{Llama-3.2-1B}
    & 0\%            & -100\%    & 38.33 & 35.53 & 33.76 & 45.71 \\
    & 100\%          & 0\%       & 39.36 & 35.94 & \textbf{36.86} & 45.29 \\
    \cline{2-7}
    & RAND           & +7.01\%   & 39.44 & 35.91 & 34.81 & \textbf{47.58} \\
    & IFD            & -15.77\%  & 39.20 & 37.35 & 34.47 & 45.78 \\
    & DEITA          & -65.79\%  & 38.69 & 36.58 & 33.45 & 46.03 \\
    & AlpaGasus      & -47.39\%  & 38.88 & 36.89 & 33.87 & 45.86 \\
    & SSPL           & +69.55\%  & 40.08 & 37.20 & 36.60 & 46.44 \\
    & \textbf{RAISE(Ours)} & \textbf{+85.14\%} & \textbf{40.24} & \textbf{38.14} & 35.58 & 47.01 \\
    \midrule
    \multirow{7}{*}{Qwen-2.5-3B}
    & 0\%            & -100\%    & 62.16 & 63.19 & 47.13 & 76.15 \\
    & 100\%          & 0\%       & 63.70 & 65.22 & 49.12 & 76.75 \\
    \cline{2-7}
    & RAND           & +15.53\%  & 63.94 & 65.30 & 50.18 & 76.33 \\
    & IFD            & +34.04\%  & 64.22 & 65.25 & 50.43 & 76.99 \\
    & DEITA          & +28.37\%  & 64.13 & \textbf{65.43} & 49.66 & 77.31 \\
    & AlpaGasus      & +32.06\%  & 64.19 & 65.18 & 49.91 & 77.48 \\
    & SSPL           & +46.59\%  & 64.42 & 65.33 & 50.68 & 77.23 \\
    & \textbf{RAISE(Ours)} & \textbf{+69.83\%} & \textbf{64.77} & 65.32 & \textbf{51.28} & \textbf{77.72} \\
    \bottomrule
    \end{tabular}
    \caption{\textbf{Performance comparison on MMLU, ARC-Challenge, and CommonsenseQA.} All methods are trained on Alpaca-52K. We report results for two versions of Llama-3.2 (3B and 1B) and Qwen-2.5-3B. ``\textbf{0\%}'' denotes the base model and ``\textbf{100\%}'' denotes full Alpaca dataset, and otherwise we select 1\% of the data or equivalent number of training steps. ``\textbf{Avg}'' denotes the average metric across these three benchmarks. ``\textbf{Avg.Q}'' denotes the additional performance gain achieved by each method, relative to the improvement obtained by using the 100\% data baseline. It is computed as: $\textbf{Avg.Q}(\cdot) = \frac{\text{Avg}(\cdot) - \text{Avg}(100\%)}{\text{Avg}(100\%) - \text{Avg}(0\%)}$. 
    Bold numbers denotes the best performing on its column.}
    \label{tab:main1}
\end{table*}

\begin{figure}
    \centering
    \includegraphics[width=1\linewidth]{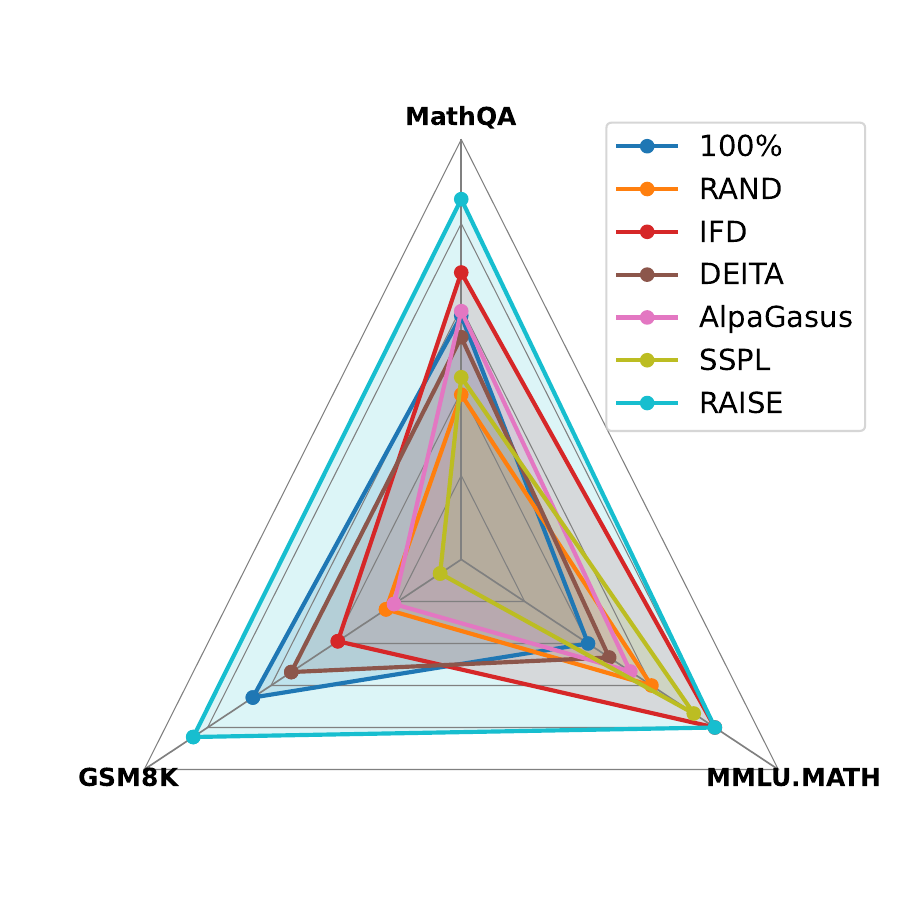}
    \caption{\textbf{Performance of GSM8K-targeted training.} All methods are still trained on Alpaca dataset but use GSM8K as validation set (different from the GSM8K evaluation set). Besides, We use MathQA and MMLU.Math (MMLU.college\_mathematics) for OOD evaluations.}
    \label{fig:gsm_dadar}
\end{figure}
We present the results of RAISE versus various baselines using different models in Tables~\ref{tab:main1}, and we showcase RAISE’s capability for task-specific optimization in Figure~\ref{fig:gsm_dadar}. Our key findings are as follows:

\paragraph{Only 1\% of gradient-update steps suffices to surpass full-data training.}
In Table~\ref{tab:main1}, RAISE requires only 1\% of the total number of update steps, yet outperforms models trained on the entire dataset on all models. Notably, RAISE achieves a significantly better result than this full-data baseline. We conjecture that only a small fraction of data truly benefits the task objective, while most of the dataset provides minimal gains. By explicitly optimizing toward the task objective, RAISE effectively captures these valuable data. In Qwen's experimental results, using the entire dataset actually yielded worse results than using a subset of the data. To some extent, this further supports this point. We conducted a more detailed analysis of this phenomenon of "less is more" in Appendix~\ref{app:less_is_more}.

\paragraph{RAISE consistently outperforms baselines on different models.}
Tables~\ref{tab:main1} shows that RAISE achieves superior performance across all tested models. Although RAISE remains robust for both small and large model scales, its advantage over baselines is especially pronounced on stronger Llama-3.2-3B compared to smaller Llama-3.2-1B.

\paragraph{RAISE exhibits strong capability of task-specific optimization.}
In Figure~\ref{fig:gsm_dadar}, all baselines perform poorly in GSM8K due to their reliance on heuristic and general ``quality'' metrics, which predominantly capture instruction difficulty rather than the actual task objective. Since only a small fraction of Alpaca’s instructions involve the target reasoning tasks, these baselines are largely ineffective. In contrast, RAISE explicitly identifies and prioritizes instructions that align with the final objective, as evidenced by its emphasis on computational and reasoning-focused prompts relevant to GSM8K (see Appendix~\ref{app:case_study}). Meanwhile, we conducted out-of-distribution evaluations on MathQA and MMLU.Math (MMLU.college\_mathematics), and RAISE still outperformed the baseline.

\section{Analysis}
In this section, we further investigate how RAISE selects instructions by examining two core modules: \textbf{state fusion} and \textbf{diversity-constrained selection}. Finally, we analyze the distribution of data selected by RAISE at different stages of training.
\subsection{Ablation on State Fusion}
\begin{figure}
    \centering
    \includegraphics[width=1\linewidth]{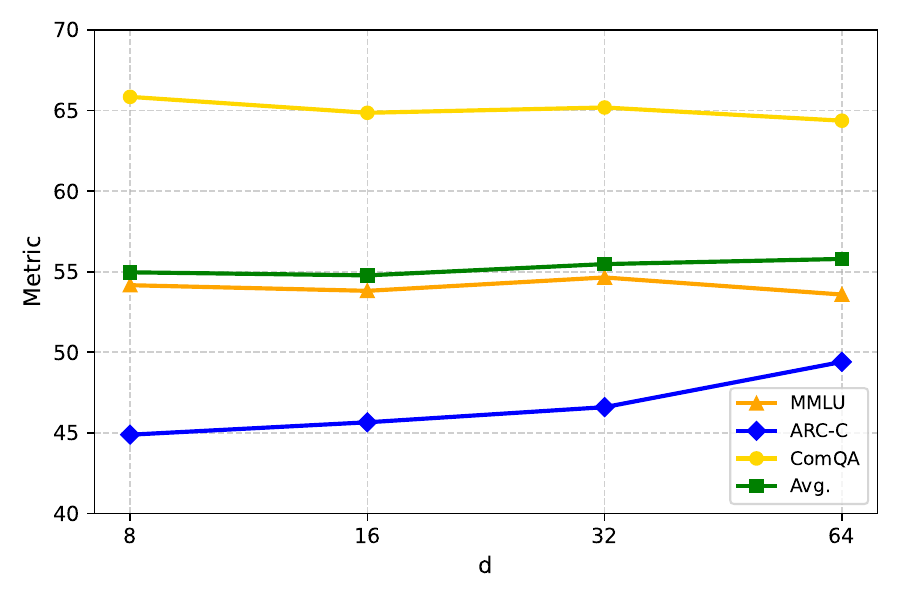}
    \caption{Performance with different instruction semantic dimensions $d_{sem}$.}
    \label{fig:abl_sem}
\end{figure}
\paragraph{A small instruction semantic dimension suffices.} We vary the dimension of the semantic embedding (\(\{8,16,32,64\}\); default: 32) and report the results in Figure~\ref{fig:abl_sem}. Even though 32 is much smaller than the original embedding size (e.g., 4096), it consistently yields solid performance on MMLU, ARC-C, and ComQA. Increasing the semantic dimension leads to modest gains overall, but notably, ARC-C benefits the most from higher-dimensional representations, suggesting a stronger reliance on richer feature spaces for reasoning. Although performance on MMLU and ComQA slightly declines at 64 dimensions, the improvement on ARC-C compensates, keeping the overall average competitive.

\paragraph{Stage State is the most critical.} 
\begin{figure}[ht]
    \centering
    \includegraphics[width=1\linewidth]{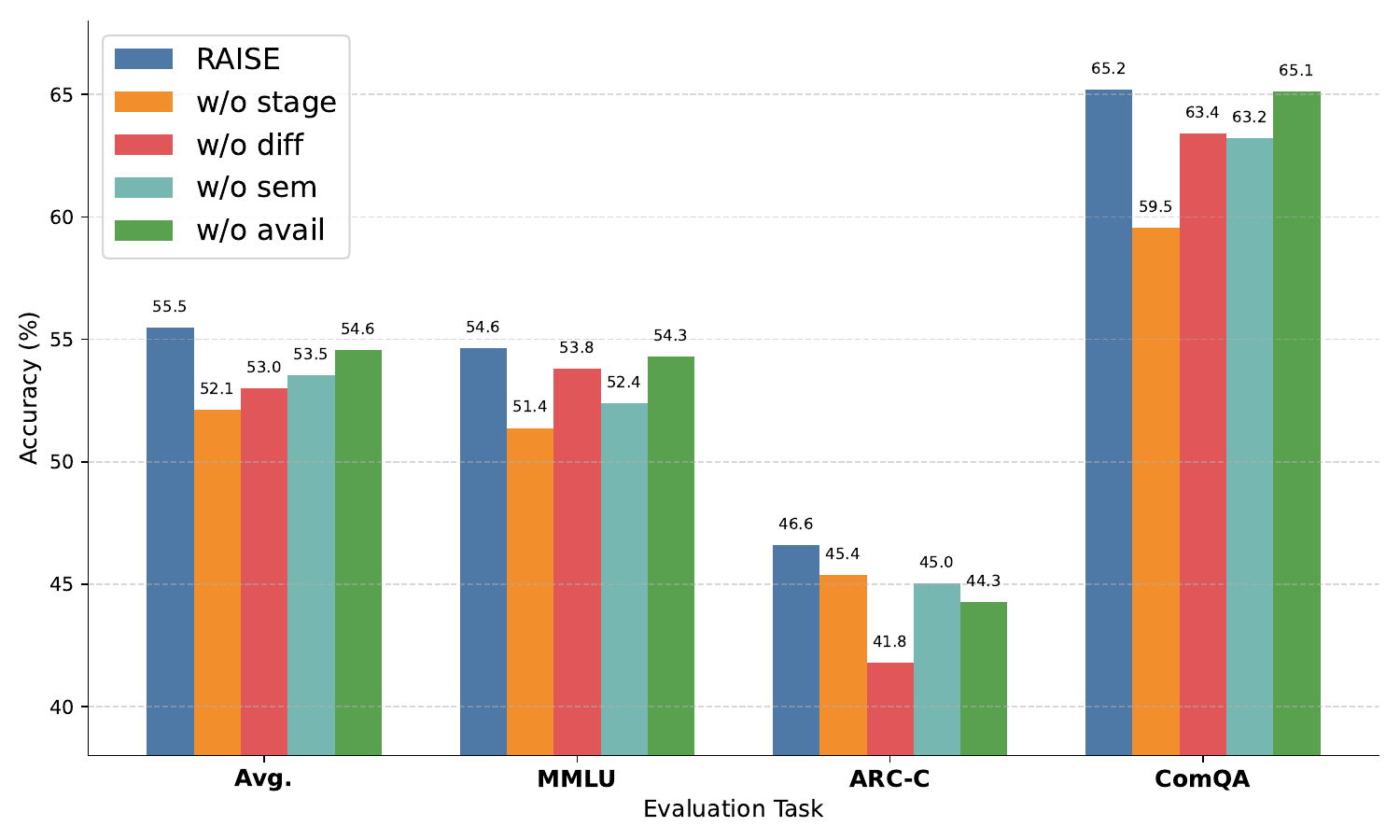}
    \caption{Ablation results on different components in \textbf{state fusion}.}
    \label{fig:abl_component}
\end{figure}
Ablation results in Figure~\ref{fig:abl_component} show that removing the Stage State leads to the largest performance drop, confirming its role as the key temporal controller in RAISE. It dynamically guides training by integrating model status and progress, enabling early semantic exploration and late-stage sample refinement. Other states (difficulty, semantic, availability) also contribute—e.g., Difficulty State is essential for complex tasks like ARC-C, and Semantic State helps in knowledge-rich domains like MMLU—but their impact is more task-specific. Availability plays a minor role but ensures diversity with minimal cost. We discuss in detail in Appendix~\ref{app:component_effect} the impact of each component on the performance of RAISE.

\begin{figure}[ht]
    \centering
    \includegraphics[width=1\linewidth]{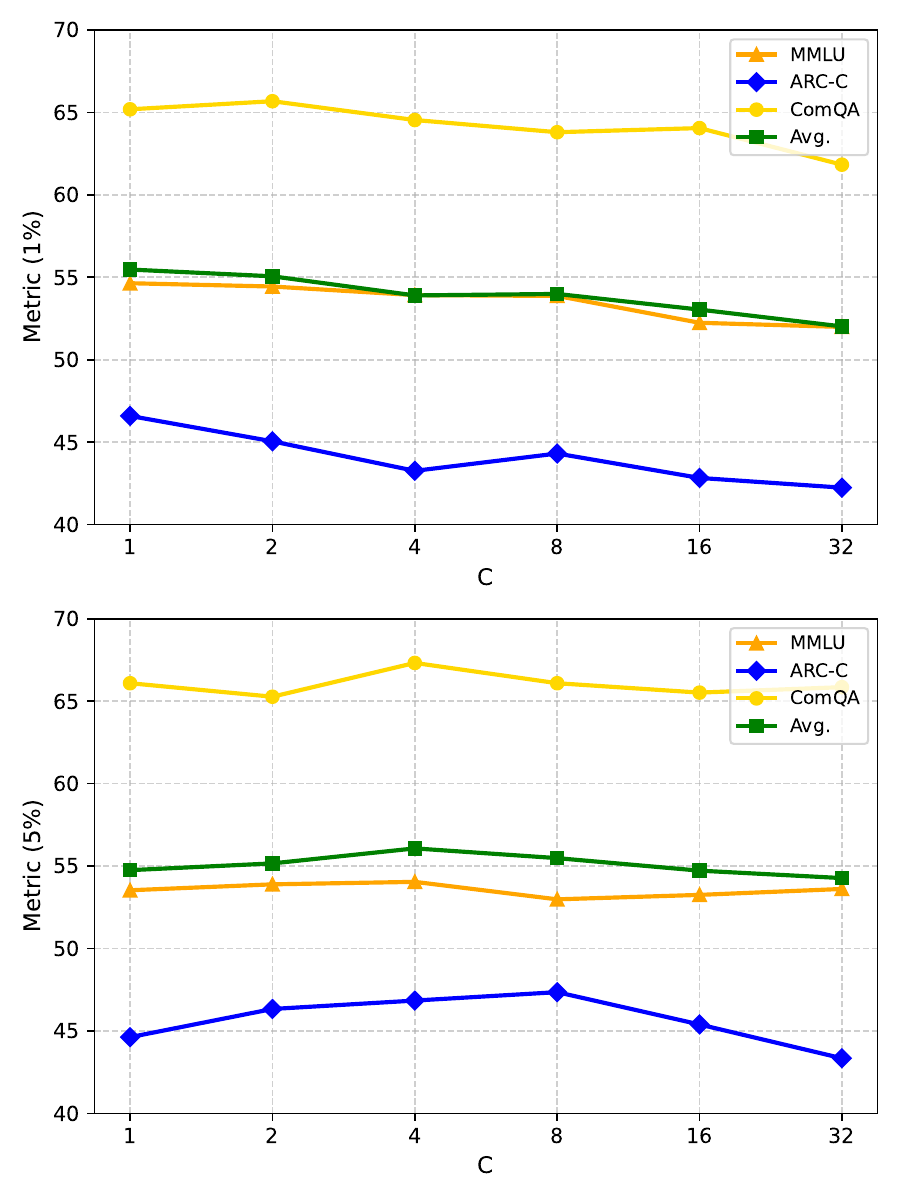}
    \caption{Performance with different class counts \(C\) when using only 1\% and 5\% data. The larger the \(C\), the finer the class, and the fewer instructions each class selects.}
    \label{fig:abl_div}
\end{figure}
\subsection{Ablation on Diversity-Constrained Selection}
In the diversity-constrained selection, all data are first clustered into \(C\) classes via K-means, and the model then selects top-scoring samples within each class. We study how different values of \(C\) affect performance. As shown in Figure~\ref{fig:abl_div}, we vary \(C \in \{1,2,4,8,16,32\}\). When \(C\) is small (1 or 2), the model achieves relatively strong overall performance when selecting 1\% data, whereas larger \(C\) leads to a downward trend. While this might seem counterintuitive—given that diversity often boosts performance -- the key factor here is that RAISE uses only 1\% of the training steps compared to full-data training. Under such a tight budget, the model must rapidly focus on data most aligned with the target objective. These valuable samples may all fall into a single class, and the diversity constraint then limits how many can be selected from that cluster (\(B/C\)), thereby hurting performance. The experiment on 5\% data proves this (achieving best performance at \(C=4\)), suggesting that ensuring data diversity remains a beneficial measure for training when there is a large amount of data samples.

\subsection{Distribution of Selected Instructions}
\begin{figure}[ht]
    \centering
    \includegraphics[width=1\linewidth]{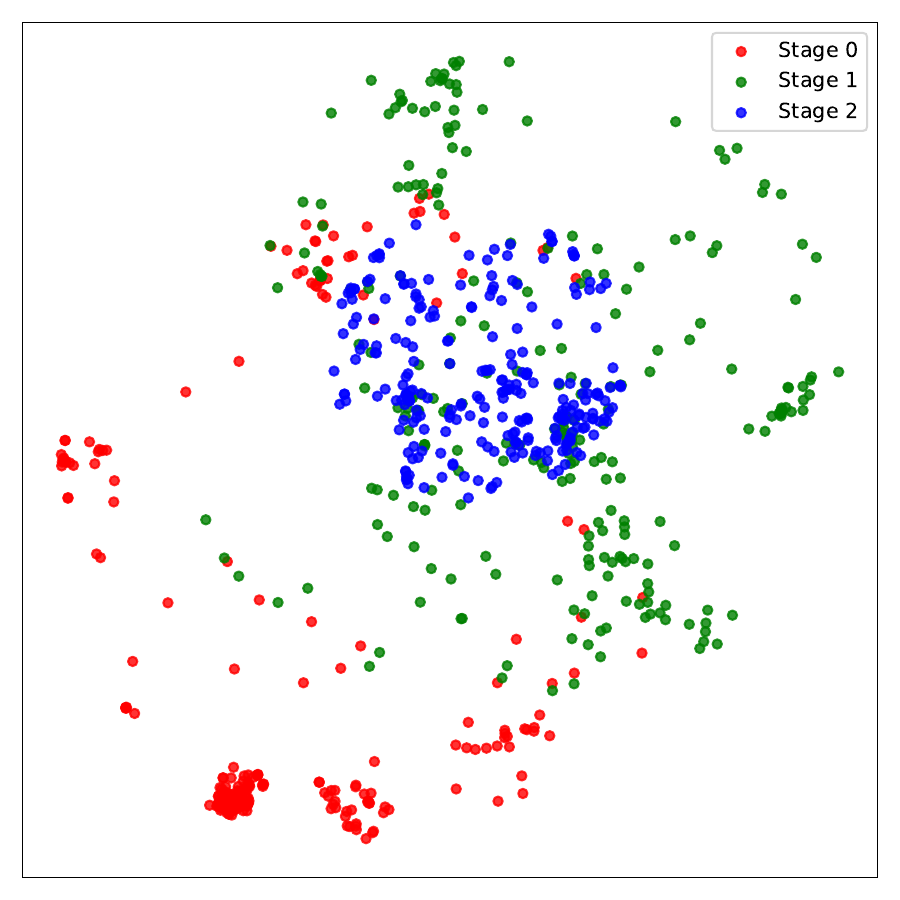}
    \caption{Distribution of selected instructions in different stage.}
    \label{fig:dist}
\end{figure}
In this section, we investigate the data selected by RAISE. We split the total $T$ training steps into three stages (Stage~0, Stage~1, and Stage~2), representing the early, middle, and late stages of training. We then visualize the distribution of the selected samples at each stage. As shown in Figure~\ref{fig:dist}, the data chosen in the early and middle stages are widely scattered, whereas in the final stage they become tightly clustered. This indicates that the most beneficial data for the model changes over time. In the early and middle stages of training, various patterns of data are helpful for the training process. However, in the later stage, the data that is most beneficial for training is concentrated on certain patterns.

\section{Conclusion}
In this paper, we present \textbf{RAISE}, a dynamic instruction selection method that adaptively selects beneficial instructions for LLM fine-tuning. RAISE employs a task-objective-driven acquisition function and a cluster-based diversity mechanism to identify high-utility data. Our experiments on multiple benchmarks demonstrate that RAISE outperforms static selection baselines, achieving strong performance while using only a small fraction of training steps. We hope this work inspires further research on adaptive data selection and fine-tuning strategies for large language models.

\section*{Limitation}
RAISE incurs linear memory overhead in the replay buffer during RL training of the acquisition function. Specifically, when storing states for RL optimization, each instruction's fused state vector (dimension $M$) requires $O(M)$ memory. For a dataset of size $N$, the total buffer storage scales as $O(N \times M)$. This becomes prohibitive for composite datasets where $N \geq 200,000$---common in current instruction tuning. Furthermore, when sampling batches from the buffer, multiple state vectors must be simultaneously loaded into memory, exacerbating peak memory pressure. Consequently, RAISE faces scalability challenges for very large-scale instruction datasets, necessitating future work on state compression or distributed buffer strategies.

\section*{Acknowledgement}
This research is supported by National Natural Science Foundation of China (Grant No.62276154), Research Center for ComputerNetwork (Shenzhen) Ministry of Education, the Natural Science Foundation of Guangdong Province (Grant No.2023A1515012914 and 440300241033100801770), Basic Research Fund of Shenzhen City (Grant No.JCYJ20210324120012033, JCYJ20240813112009013 and GJHZ20240218113603006), the Major Key Projectof PCL for Experiments and Applications (PCL2023A09). This work is also supported in part by NSF under grants III-2106758, and POSE-2346158.

\bibliography{main}
\newpage
\appendix

\section{Details of Experiments Setup}
\subsection{Evaluation Datasets}
\label{app:eval_detail}
MMLU covers 57 tasks ranging from elementary math and U.S. history to computer science and law, primarily measuring knowledge breadth and reasoning. ARC-C is a challenging subset of the AI2 Reasoning Challenge, featuring multiple-choice questions that demand complex reasoning and scientific knowledge \cite{allenai:arc}. ComQA is a common-sense reasoning benchmark requiring real-world knowledge and inference \cite{talmor-etal-2019-commonsenseqa}. GSM8K contains 8,000 grade-school math problems focusing on multi-step numeric reasoning \cite{cobbe2021gsm8k}. Table~\ref{tab:evaluation_dataset} provides more detailed information about these evaluation datasets.

\begin{table}[ht] 
\centering
\small 
\renewcommand{\arraystretch}{1.2} 
\begin{tabular}{@{}llccc@{}}
\toprule
\textbf{Dataset} & \textbf{|$\mathcal{D}_{\text{val}}$|} & \textbf{|$\mathcal{D}_{\text{test}}$|} & \textbf{Answer Type} & \textbf{Metric} \\
\midrule
MMLU   & 285                                & 14,042                              & Letter options       & Acc             \\
ARC-C  & 299                                  & 1,172                               & Letter options       & Acc             \\
ComQA  & 280 \tablefootnote{In ComQA, we randomly select 280 samples from the original 1,221 validation data.}                                 & 1,140                               & Letter options       & Acc             \\
GSM8K  & 256 \tablefootnote{For GSM8K, which does not have a dedicated validation set, we sample 256 examples from its 7,473 training data.}                                  & 1,319                               & COT and answer       & Acc             \\
\bottomrule
\end{tabular}
\caption{Statistics of evaluation datasets.}
\label{tab:evaluation_dataset}
\end{table}

\subsection{Hyperparameters}
When training the LLM in each round, we set the learning rate to 2e-5, use a cosine learning rate scheduler, and have no warm-up steps. When ppo training, we set actor learning rate, critic learning rate, weight decay, $\gamma$, $\lambda$, K respectively to 1e-1, 2e-1, 1e-2, 0.99, 1.0, 20. In our \textbf{state fusion} pipeline, we pool the instruction embedding vector to a dimension of 32. Consequently, the fused state dimension is \(2\) (stage) + \(4\) (diff) + \(32\) (sem) + \(1\) (avail) = \(39\). As a practical measure of model performance \(\mathcal{P}(\mathcal{M}_t)\), we use \(-\mathrm{Loss}(\mathcal{M}_t, \mathcal{D}_{\text{val}})\) for computational efficiency. When evaluating the trained LLM, we use the lm\_eval framework and set max\_length to 512.

\subsection{SSPL Baseline}
\label{app:sspl}
In SSPL baseline, all training examples are sorted by their loss values and divided into max\_steps buckets of approximately equal size, such that each bucket contains instructions with similar difficulty (as measured by loss). During training, the model sequentially takes data batches from these buckets in ascending order of difficulty, moving from simpler to more challenging tasks to progressively enhance its capabilities.

\section{Effects of Different Components in State Fusion}
\label{app:component_effect}
The ablation experimental results in Figure~\ref{fig:abl_component} systematically evaluate the 4 state components, revealing the following importance ranking: \textbf{Stage} > \textbf{Instruction Difficulty} > \textbf{Instruction Semantic} > \textbf{Instruction Availability}. We provide a more detailed analysis here.
\subsection{Stage State}
As the core component of RAISE's dynamic selection, Stage State integrates current model performance and training progress to provide global temporal awareness. Figure~\ref{fig:abl_component} results show removing this component causes the largest average performance drop, as it directly controls training rhythm: guiding broad semantic category exploration in early stages (dispersed distributions in early/mid phases in Figure~\ref{fig:dist}) and focusing on high-value samples later (dense distributions in later phases). Its global and dynamic nature makes it most significant in multi-task scenarios, serving as the key source of temporal sensitivity for dynamic value estimation.
\subsection{Instruction Difficulty State}
Drives progressive learning by quantifying instruction complexity. Ablation experiments show significant performance drops on ARC-Challenge when removed, as this task heavily relies on complex reasoning (e.g., scientific knowledge inference). Works synergistically with Stage State: selecting easier samples early for stable training, then gradually introducing harder ones. However, in knowledge-intensive tasks (e.g., MMLU), difficulty metrics show weaker correlation with task objectives, making its impact relatively limited. This indicates Difficulty State's effectiveness varies significantly by task type, being crucial for complex reasoning tasks.
\subsection{Instruction Semantic State}
Captures semantic relevance through embedding vectors. Ablation shows significant impact on MMLU and CommonsenseQA but limited effect on ARC-Challenge. This difference stems from varying task requirements: MMLU needs cross-domain knowledge generalization where semantic features prevent bias toward frequent domains (e.g., law, medicine). CommonsenseQA requires connecting similar common sense questions (e.g., causal reasoning) to extract patterns. ARC-Challenge's scientific reasoning depends more on logical chain completeness than semantic discrimination. Thus, Semantic State offers limited benefits for reasoning tasks.
\subsection{Instruction Availability State}
Tracks selection frequency to prevent oversampling. Ablation shows minimal impact, with analysis revealing <10\% cases of repeated sampling (>3 times) of same data. Its function becomes weaker with higher cluster counts (C). However, since it adds negligible overhead (just 1 extra dimension) and handles extreme oversampling cases, retaining it remains meaningful.

\section{Dynamic Value}
\label{app:dynamic_value}

\subsection{Static Value Framework}
Existing approaches rely on \textit{static value} computation. Let $f: d \rightarrow \mathbb{R}$ be the scoring function where $\forall d \in \mathcal{D}$, $f(d)$ represents its general utility. The subset selection objective is:
\begin{align}
    \mathcal{D}' &= \mathop{\mathrm{arg\,top}}_{N'} \big\{ f(d) \mid d \in \mathcal{D} \big\} \\
    \mathcal{J} &= \max_f \mathcal{P}(\mathcal{M}[\mathcal{D}'])
\end{align}

where $N'$ is the selection size and $\mathcal{P}$ denotes evaluation metrics. Representative implementations include:
\begin{itemize}
    \item \textbf{AlpaGasus} \cite{chen2023alpagasus}: $f_{\text{GPT-4}}(d) = \text{Complexity}(d) + \text{FormatScore}(d)$
    \item \textbf{IFD} \cite{li-etal-2024-quantity}: $f_{\text{Llama}}(d) = \frac{\log p(y|x)}{\log p(y)}$
\end{itemize}

\subsection{Dynamic Value Definition}
We extend static values to \textit{dynamic values} by introducing temporal dependency:
\begin{equation}
    f: \mathcal{M}_{t-1} \times \mathcal{D} \rightarrow \mathbb{R}, \quad \text{where } t \in \{1,\dots,T\}
\end{equation}
The training dynamics are governed by:
\begin{equation}
    \mathcal{D}_t = \mathop{\mathrm{arg\,top}}_{B} \big\{ f_t(\mathcal{M}_{t-1}, d) \mid d \in \mathcal{D} \big\}
    \label{eq:selector}
\end{equation}
\begin{equation}
    \mathcal{M}_t = \mathrm{Update}(\mathcal{M}_{t-1}, \mathcal{D}_t)
    \label{eq:updater}
\end{equation}
with final objective:
\begin{equation}
    \mathcal{J} = \max_{f} \mathcal{P}(\mathcal{M}_T)
\end{equation}

\subsection{Conceptual Comparisons}
\begin{itemize}
    \item \textbf{Vs. Reward}: Reward $R_t$ measures immediate performance improvement $\Delta\mathcal{P}_t$, whereas $f(\mathcal{M}_{t-1},\mathcal{D})$ estimates long-term utility of each sample.
    \item \textbf{Vs. Curriculum Learning}: Dynamic values adapt to emergent model states rather than pre-defined difficulty schedules.
    \item \textbf{Vs. Active Learning}: Focuses on final model capability rather than immediate uncertainty reduction.
\end{itemize}

\subsection{Example}
Consider $\mathcal{D} = \{d_1, d_2\}$ with $B=1$ at step $t$:
\begin{align*}
    f(\mathcal{M}_{t-2}, d_1) &= 0.8 \\
    f(\mathcal{M}_{t-2}, d_2) &= 0.1 \\
    \Rightarrow \mathcal{D}_t &= \{d_1\} \quad (\text{via Eq.~\ref{eq:selector}})
\end{align*}
The model updates as $\mathcal{M}_{t-1} \rightarrow \mathcal{M}_t$ following Eq.~\ref{eq:updater}. The value predictor $f$ (we call it \textbf{acquisition function}) is optimized via RL. 

\subsection{Parameter-to-Time Simplification}
Directly using model parameters in $f(\mathcal{M}_t,d)$ is computationally prohibitive. We simplify to $f(t,d)$ because optimizers like Adam \cite{kingma2014adam} induce smooth parameter updates, making $t$ a sufficient temporal index.
The simplification reduces complexity from $\mathcal{O}(|P|\times |\mathcal{D}|)$ to $\mathcal{O}(|\mathcal{D}|)$, enabling real-time computation.

\begin{algorithm}
\small
\caption{Selection Policy Optimization}
\label{alg:ppo}
\begin{algorithmic}[1]
\State \textbf{Input:} Training dataset \( \mathcal{D} \), Validation set \( \mathcal{D}_{\text{val}} \), Initial LLM \( \mathcal{M}_0 \), Number of rounds \( K \), Steps per round \( T \), Batch size \( B \), Fusion function \( \mathcal{H} \), PPO epochs \( K_2 \)
\State \textbf{Output:} \( \pi_{\theta_{K}} \)
\State Initialize \( \theta_0 \), \( \phi_0 \)
\For{\( k = 1 \) \textbf{to} \( K \)}
    \State // Data Collection Phase
    \State Initialize buffer \(W \leftarrow \emptyset \)
    \For{\( t = 1 \) \textbf{to} \( T \)}
        \State \( S_t \gets \mathcal{H}(\mathcal{D}, \mathcal{M}_{t-1}, t) \)
        \State \( \mathcal{D}_t \gets \pi_{\theta_{k-1}}(S_t, B) \)
        \State \( V_t \gets V_{\phi_{k-1}}(S_t) \)
        \State \( \mathcal{M}_t \gets \text{Update}(\mathcal{M}_{t-1}, \mathcal{D}_t) \)
        \State \( R_t \gets \mathcal{P}(\mathcal{M}_t)-\mathcal{P}(\mathcal{M}_{t-1}) \)
        \State \(W \leftarrow W \cup \{(S_t, \mathcal{D}_t, R_t, V_t, S_{t+1})\} \) 
    \EndFor
    \State // Policy Optimization Phase
    \State \( \theta, \theta_\text{old}, \psi = \theta_{k-1}, \theta_{k-1}, \psi_{k-1} \)
    \For{\( k_2 = 1 \) \textbf{to} \( K_2 \)}
        \For{\( (S_t', \mathcal{D}_t', R_t', V_t', S_{t+1}') \subseteq W \)}
            \State Update \( \theta \) by Eq.\ref{loss_actor}
            \State Update \( \psi \) by Eq.\ref{loss_critic}
        \EndFor
    \EndFor
    \State \( \theta_k, \psi_k = \theta, \psi \)
\EndFor    
\State \textbf{return} \( \pi_{\theta_K} \)
\end{algorithmic}
\end{algorithm}
\section{Selection Policy Optimization}
\label{app:ppo}
The optimization algorithm for the data selection strategy (acquisition function) is illustrated in Algorithm~\ref{alg:ppo}. This framework employs a bi-level iterative optimization process:
\begin{itemize}
\item The inner loop trains the LLM and adaptively updates the data selection strategy based on observed performance improvements during training.
\item The outer loop iteratively refines the selection strategy over multiple rounds, ultimately achieving global optimization of the policy $\pi_\theta$.
\end{itemize}

\section{Analysis of Computational Cost}
RAISE introduces moderate additional overhead due to its two-stage structure: instruction preprocessing and acquisition function training. We denote $\Phi$ as the cost of full-data LLM training.
\paragraph{Instruction Preprocessing.} This stage computes instruction difficulty and semantic representations using an auxiliary model. Each instruction requires two forward passes. Assuming the auxiliary model is 3$\times$ larger than the target model, and training is 5$\times$ more expensive than inference per epoch, the preprocessing cost is:
\[
\text{Preprocess} \approx \frac{2 \times 3}{5 \times 3} \Phi = 0.4\Phi
\]

\paragraph{Acquisition Function Training.} The acquisition function is trained via a bi-loop structure over $K$ rounds. Each round performs one LLM training using a $p$-proportion data subset, costing $p\Phi$. Additional PPO and selection overheads are negligible ($<80$s per round). Total training cost is:
\[
\text{Train} \approx K \cdot p \cdot \Phi
\]
Using $p=0.05$, $K=30$, we obtain $1.5\Phi$.

\paragraph{Total Cost.} The total computational cost of RAISE is approximately:
\[
\text{Total} \approx 0.4\Phi + 1.5\Phi = 1.9\Phi
\]

Despite a 90\% increase over full-data training, RAISE enables targeted instruction selection and facilitates constructing high-quality, task-aligned datasets. For instance, aligning Alpaca with GSM8K as validation allows extraction of reasoning-specific data for specialized downstream training.

\section{Memory Optimization Strategy}
During RL training of acquisition function, RAISE maintains a replay buffer to store fused state vectors for each instruction. Given a state dimension of $M$ and dataset size $N$, the memory cost scales linearly as $O(N \times M)$. This becomes prohibitive for large-scale instruction datasets (e.g., $N \geq 200{,}000$), where simultaneous access to multiple state vectors during batch sampling can exacerbate peak memory pressure.

To address this scalability issue, we propose a simple strategy: performing data selection every $M$ steps instead of every training step. During intermediate steps, random sampling is used in place of selection. This reduces the frequency of policy updates and state maintenance, decreasing both computational and memory costs to approximately $1/M$ of the original per-step selection setup.

This optimization leverages the observation that model parameters typically change slowly between adjacent training steps. Thus, the stage state remains similar over short horizons, and infrequent updates (e.g., every $M$ steps) preserve most of the benefits of dynamic selection while significantly reducing overhead.

The interval parameter $M$ provides a tunable trade-off between adaptivity and efficiency:

\begin{itemize}
    \item \textbf{Fully Dynamic Selection ($M=1$):} Per-step selection with maximum adaptivity but highest memory and compute cost.
    \item \textbf{Fully Static Selection ($M=N$):} Single selection before training with no runtime adaptation and minimal overhead.
\end{itemize}

This parameterization allows explicit control over the adaptation-efficiency trade-off in practical deployments. For very large-scale instruction tuning, this strategy enables RAISE to scale more efficiently without significant loss in performance.

\section{Additional experiments}
\begin{table*}[htb]
    \centering
    \begin{tabular}{c c c c| c c c c}
    \toprule
    \textbf{Model} & \textbf{DATA} & \textbf{Avg.Q} & \textbf{Avg.} & \textbf{MMLU} & \textbf{ARC-C} & \textbf{ComQA} & \textbf{GSM8K} \\
    \midrule
    \multirow{8}{*}{Llama-3.1-8B}
    & 0\%            & -100\%    & 57.12 & 63.35 & 51.11 & 71.42 & 42.61 \\
    & \textbf{100\%}          & \textbf{0\%}       & \textbf{61.48} & 63.47 & \textbf{55.29} & \textbf{72.48} & \textbf{54.66} \\
    \cline{2-8}
    & RAND 1\%       & -72.84\%  & 58.30 & 63.47 & 53.67 & 71.50 & 44.58 \\
    & IFD 1\%        & -91.42\%	 & 57.49 & 63.46 & 52.62 & 69.86 & 44.03 \\
    & DEITA 1\%      & -80.25\%	 & 57.98 & \textbf{64.19} & 52.73 & 71.25 & 43.75 \\
    & AlpaGasus 1\%  & -71.92\%	 & 58.34 & 63.79 & 53.58 & 71.42 & 44.58 \\
    & LESS 1\%       & -51.49\%	 & 59.23 & 64.02 & 53.33 & 71.74 & 47.84 \\
    & RAISE 1\% & -40.99\%	 & 59.69 & 64.17 & 54.69 & 71.33 & 48.56 \\
    \midrule
    \multirow{8}{*}{Llama-3.1-8B}
    & 0\%            & -100\%    & 57.12 & 63.35 & 51.11 & 71.42 & 42.61 \\
    & 100\%          & 0\%       & 61.48 & 63.47 & 55.29 & 72.48 & 54.66 \\
    \cline{2-8}
    & RAND 5\%       & -36.97\%	 & 59.87 & 64.02 & 53.41 & 72.32 & 49.71 \\
    & IFD 5\%        & -45.10\%	 & 59.51 & 63.15 & 53.33 & 69.78 & 51.79 \\
    & DEITA 5\%      & -18.13\%	 & 60.69 & 64.02 & 53.41 & 72.32 & 52.99 \\
    & AlpaGasus 5\%  & -38.06\%	 & 59.82 & 63.60 & 56.40 & 72.73 & 46.55 \\
    & LESS 5\%       & +12.99\%	 & 62.04 & \textbf{64.29} & 55.12 & \textbf{72.81} & 55.95 \\
    & RAISE 5\% & \textbf{+29.55\%}	 & \textbf{62.76} & 64.09 & \textbf{55.97} & 72.24 & \textbf{58.76} \\
    \bottomrule
    \end{tabular}
    \caption{Experimental results of the data from 1\% and 5\% of Tulu3.}
    \label{tab:llama3_8B_tulu3}
\end{table*}
To further verify the scalability and effectiveness of our method, we conducted supplementary experiments on the larger Llama-3.1-8B model and the higher-quality Tulu3 dataset, and included a stronger baseline: LESS \cite{xia2024less}. The results are reported in Table~\ref{tab:llama3_8B_tulu3}. As shown in the table, RAISE still outperforms LESS and other baselines at both 1\% and 5\% budgets. However, we observe that on the high-quality Tulu3 dataset, no method can surpass full-data training when only 1\% of the data are used. Only when the budget increases to 5\% do LESS and RAISE marginally exceed the performance of the 100\% baseline. This suggests that the ``Less is More'' phenomenon, while pronounced on lower-quality data, is less dramatic on high-quality data where the intrinsic value density of every sample is already high.

\section{Why ``Less Is More''?}
\label{app:less_is_more}
We observe that full-data training incurs significantly higher computational cost—up to $100\times$ compared to selective training baselines—while resulting in comparable or only marginally improved performance. This finding may appear to contradict existing scaling laws. However, as noted in prior work~\cite{zhou2024lima}, scaling laws do not directly apply to the alignment phase, for the following reasons:
\begin{itemize}
    \item The majority of LLMs' core capabilities are acquired during pretraining.
    \item The alignment phase primarily teaches output patterns that activate these pretrained capabilities.
    \item Thus, only a small amount of high-quality data is sufficient for effective alignment.
\end{itemize}

\begin{figure}[ht]
    \centering
    \includegraphics[width=1\linewidth]{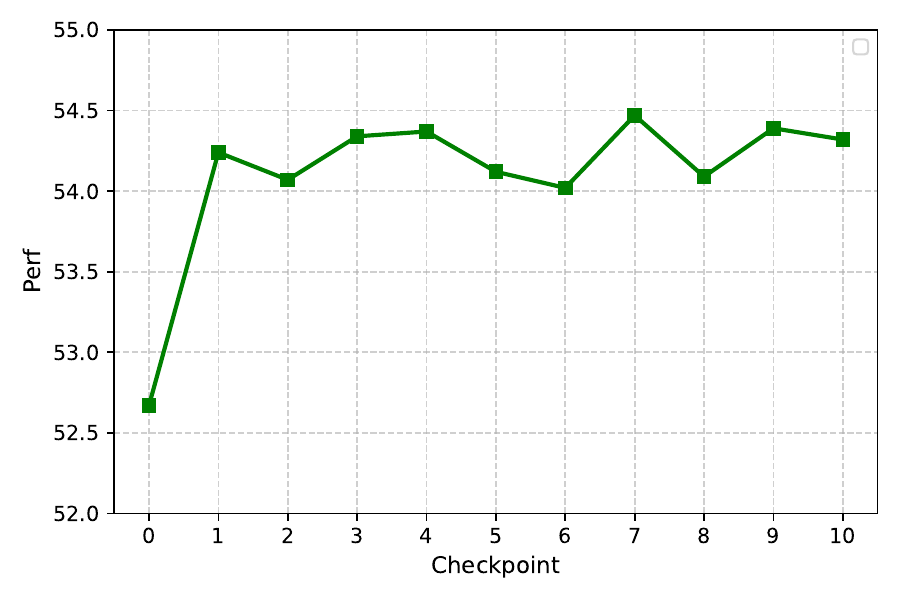}
    \caption{Model performance during training.}
    \label{fig:lima}
\end{figure}
\paragraph{Checkpoint Analysis.} To further examine this behavior, we conducted full-data training and saved 10 equally spaced checkpoints. Figure~\ref{fig:lima} reports performance across checkpoints. The results show that performance plateaus after early training steps, with only minor fluctuations throughout the rest of training. This reinforces the hypothesis that only a small subset of the full data contributes meaningfully to alignment.

\section{Bias of Auxiliary Model}
\begin{figure}[htbp]
    \centering
    \includegraphics[width=1\linewidth]{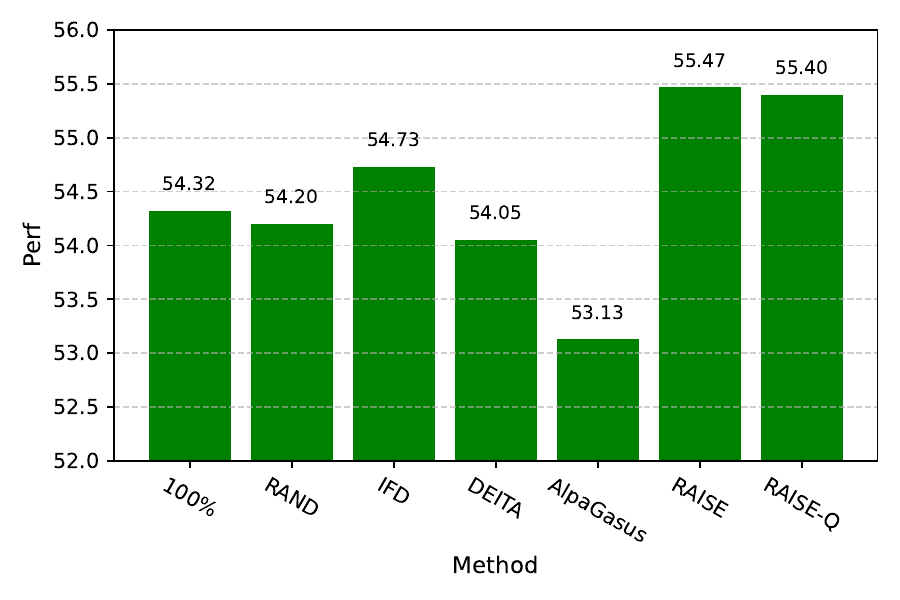}
    \caption{Experiment result of using Qwen-2.5-7B-Instruct as auxiliary model (RAISE-Q).}
    \label{fig:abl_aux_bias}
\end{figure}
The Instruction Difficulty and Instruction Semantic features computed by auxiliary models are crucial for RAISE's performance. We emphasize these components exhibit strong generalizability, while absolute values may differ significantly between architecturally distinct models (e.g., GPT vs. Llama), their feature distributions remain similar. To validate this experimentally, we implemented RAISE-Q using Qwen-2.5-7B-Instruct as the auxiliary model. The experimental results are shown in Figure~\ref{fig:abl_aux_bias}. It can be observed that even when using an auxiliary model of another architecture, RAISE still exhibits excellent performance, indicating that RAISE's dependence on the reference model is not particularly strong.

\section{Case Study}
\label{app:case_study}
In the experiments where GSM8K was used as the task objective, we were surprised to find that the selected instruction data were typical CoT data, which were highly aligned with GSM8K. Here we present two data cases for reference.
\begin{tcolorbox}[breakable,title=Solve for the inverse of the matrix,colframe=black]
\textbf{\#\#\# Instruction}: \\
Find the inverse of the matrix. \\
\\
\textbf{\#\#\# Input}: \\
$\begin{bmatrix}1 & -3 & 2\\3 & 2 & 1 \\ 4 & 0 & -1\end{bmatrix}$ \\
\\
\\
\textbf{\#\#\# Response}: \\
We can find the inverse of the input matrix using the adjugate or classical adjoint method. Here are the steps:

\textbf{Solution}:
First, we need to find the matrix of minors. The matrix of minors is a matrix where each element is the determinant of the minor for the corresponding entry in the original matrix. The minor for an entry is the determinant of the matrix that we get by removing the entry's row and column. 

The matrix of minors of $\begin{bmatrix}1 & -3 & 2\\3 & 2 & 1 \\ 4 & 0 & -1\end{bmatrix}$ is
$\begin{bmatrix}\begin{vmatrix}2 & 1\\0 & -1\end{vmatrix} & \begin{vmatrix}3 & 1\\4 & -1\end{vmatrix} & \begin{vmatrix}3 & 2\\4 & 0\end{vmatrix}\\ \begin{vmatrix}-3 & 2\\ 0 & -1\end{vmatrix} & \begin{vmatrix}1 & 2\\4 & -1\end{vmatrix} & \begin{vmatrix}1 & -3\\4 & 0\end{vmatrix} \\ \begin{vmatrix}-3 & 2\\2 & 1\end{vmatrix} & \begin{vmatrix}1 & 2\\3 & 1\end{vmatrix} & \begin{vmatrix}1 & -3\\3 & 2\end{vmatrix} \end{bmatrix} 
= \begin{bmatrix}-2 & -1 & -8\\3 & -6 & -12\\ -7 & -1 & 11\end{bmatrix}$.

Then we need to turn this matrix of minors into a matrix of cofactors by negating the elements at the positions where the sum of the row number and the column number is odd, i.e, alternate signs in the matrix of minors.

The matrix of cofactors is $\begin{bmatrix}-2 & 1 & -8\\-3 & -6 & 12\\ -7 & 1 & 11\end{bmatrix}$.

Next, we have to find the adjugate of the matrix is by taking the transpose of the matrix of cofactors.
\end{tcolorbox}

\begin{tcolorbox}[breakable,title=Solve the Eight Queens puzzle,colframe=black]
\textbf{\#\#\# Instruction}: \\
Design an algorithm to solve the Eight Queens puzzle \\
\\
\textbf{\#\#\# Response}: \\
The Eight Queens puzzle is a classic problem where the goal is to place eight queens on an 8x8 chessboard in such a way that no two queens threaten each other. This means that no two queens can be placed in the same row, column, or diagonal.

One solution to this problem is to use backtracking. Here is an algorithm to solve the Eight Queens puzzle using backtracking:

1. Start with an empty 8x8 chessboard and add the first queen to the first row in the first column.

2. Move to the next row and try placing a queen in each column. Check if the position is safe by verifying if any of the previously placed queens is attacking the current position. If it’s a safe position, place a queen in that column.

3.Boolean function:
\begin{lstlisting}
def is_safe(board, row, col):
    for i in range(col):
        # Check row and columns
        if board[row][i] == 1:
            return False
        # Check diagonals
        for i, j in zip(range(row, -1, -1), range(col, -1 ,-1))
            if board[i][j] == 1:
                return False
        for i, j in zip(range(row, N, 1), range(col, -1, -1))
            if board[i][j] == 1
                return False
        return True
\end{lstlisting}

4. If a queen can be placed, move to the next row and repeat step 2. If no safe position is found for the current row, go back to the previous row and move the queen to the next safe position.

5. If all eight queens have been placed on the chessboard without attacking each other, the solution is found. Print the solution.

5. Do this for rows from 1-7 since the first row already has a queen.

6. Continue the process until all possible solutions are found, or if the solutions are exhausted, and none is found.
\end{tcolorbox}

\begin{tcolorbox}[breakable,title=Solve a Tower of Hanoi problem,colframe=black]
\textbf{\#\#\# Instruction}: \\
Create an algorithm for solving a Tower of Hanoi problem \\
\\
\textbf{\#\#\# Response}: \\
The Tower of Hanoi is a classic puzzle game with a goal of moving a stack of disks from one peg to another with the following three constraints:

1. Only one disk can be moved at a time.
2. Each move consists of taking the upper disk from one peg and moving it to another peg.
3. No disk may be placed on top of a smaller disk.

Given these constraints, we can use a recursive algorithm to solve the Tower of Hanoi puzzle.

Let's define the function `towerOfHanoi(n, source, destination, auxiliary)` where `n` is the number of disks, `source` is the peg where the disks are initially stacked, `destination` is the peg where the disks need to be transferred to and `auxiliary` is the third peg that can be used as a temporary peg for moving the disks.

1. If `n == 1`, then move the top disk from `source` to `destination` peg.
2. If `n > 1`, then:
   a. Recursively call `towerOfHanoi(n-1, source, auxiliary, destination)` to move the top `n-1` disks from the `source` peg to the `auxiliary` peg using the `destination` peg as the temporary peg.
   b. Move the `n`th disk from the `source` peg to the `destination` peg.
   c. Recursively call `towerOfHanoi(n-1, auxiliary, destination, source)` to move the `n-1` disks from the `auxiliary` peg to the `destination` peg using the `source` peg as the temporary peg.
   
Below is sample code for the algorithm in Python:

\begin{lstlisting}
def towerOfHanoi(n, source, destination, auxiliary):
    if n == 1:
        print(f'Move disk 1 from {source} to {destination}')
    else:
        towerOfHanoi(n-1, source, auxiliary, destination)
        print(f'Move disk {n} from {source} to {destination}')
        towerOfHanoi(n-1, auxiliary, destination, source)
\end{lstlisting}
When the function is called with the appropriate initial values, it will print out the step-by-step solution to the Tower of Hanoi puzzle. For example:
\begin{lstlisting}
towerOfHanoi(3, 'A', 'B', 'C')
\end{lstlisting}

\end{tcolorbox}

\end{document}